\documentclass[sigconf,natbib=False]{acmart}
\AtBeginDocument{%
  }

\copyrightyear{2026}
\acmYear{2026}
\setcopyright{cc}
\setcctype{by-nc-nd}
\acmConference[SAC '26]{The 41st ACM/SIGAPP Symposium on Applied Computing}{March 23--27, 2026}{Thessaloniki, Greece}
\acmBooktitle{The 41st ACM/SIGAPP Symposium on Applied Computing (SAC '26), March 23--27, 2026, Thessaloniki, Greece}
\acmPrice{}
\acmDOI{10.1145/3748522.3779769}
\acmISBN{979-8-4007-2294-3/2026/03}

\RequirePackage[
  datamodel=acmdatamodel,
  style=acmnumeric,
  ]{biblatex}

\usepackage{hyperref}
\usepackage{url}
\usepackage{booktabs}       
\usepackage{amsfonts}       
\usepackage{nicefrac}       
\usepackage{microtype}      
\usepackage{xcolor}         
\usepackage{amsmath}
\usepackage[ruled,vlined]{algorithm2e}
\usepackage{qcircuit} 
\usepackage{algorithm2e}
\usepackage{graphicx}
\usepackage{adjustbox}
\usepackage{caption}
\usepackage{subcaption}
\usepackage{booktabs}
\begin{document}

\title{Variational Quantum Rainbow Deep Q-Network for\\Optimizing Resource Allocation Problem}
\renewcommand{\shorttitle}{Variational Quantum Rainbow Deep Q-Network for Optimizing Resource Allocation Problem}

\author{Truong Thanh Hung Nguyen}
\affiliation{%
\institution{Analytics Everywhere Lab, \\University of New Brunswick}
\city{Fredericton}
\country{Canada}}
\email{hung.ntt@unb.ca}
\orcid{0000-0002-6750-9536}

\author{Truong Thinh Nguyen}
\affiliation{%
  \institution{University of Science and Technology of Hanoi}
  \city{Hanoi}
  \country{Vietnam}}
\email{thinhnt2410944@usth.edu.vn}
\orcid{0009-0006-6248-3783}

\author{Hung Cao}
\affiliation{%
\institution{Analytics Everywhere Lab, \\University of New Brunswick}
\city{Fredericton}
\country{Canada}}
\email{hcao3@unb.ca}
\orcid{0000-0002-0788-4377}







\begin{abstract} 
Resource allocation remains NP-hard due to combinatorial complexity. While deep reinforcement learning (DRL) methods, such as the Rainbow Deep Q-Network (DQN), improve scalability through prioritized replay and distributional heads, classical function approximators limit their representational power. We introduce \textit{Variational Quantum Rainbow DQN} (VQR-DQN), which integrates ring-topology variational quantum circuits with Rainbow DQN to leverage quantum superposition and entanglement. We frame the human resource allocation problem (HRAP) as a Markov decision process (MDP) with combinatorial action spaces based on officer capabilities, event schedules, and transition times. On four HRAP benchmarks, VQR-DQN achieves 26.8\% normalized makespan reduction versus random baselines and outperforms Double DQN and classical Rainbow DQN by 4.9-13.4\%. These gains align with theoretical connections between circuit expressibility, entanglement, and policy quality, demonstrating the potential of quantum-enhanced DRL for large-scale resource allocation. Our implementation is available at: \url{https://github.com/Analytics-Everywhere-Lab/qtrl/}.
\end{abstract}



\maketitle

\section{Introduction}
\label{sec:intro}
Resource allocation represents a fundamental NP-hard combinatorial optimization problem with diverse applications across software systems, including underwater resource management~\cite{zhang2021udarmf,azghadi2024energy}, human resource allocation \cite{nguyen2021can,nguyen2024temporal}, inventory allocation \cite{ta2024solving,tan2024optimization}, and network resource distribution \cite{jeremiah2024digital,anjum2024machine}. Various solution approaches have been developed due to the problem's complexity \cite{paduraru2021task,du2023automatic}. While exact classical methods succeed for small problem instances, real-world software engineering scenarios often involve high-dimensional optimization spaces where classical approaches encounter exponential time complexity, creating a critical performance bottleneck. This computational barrier has positioned quantum computing as a promising paradigm for addressing resource allocation challenges in modern software systems, leveraging quantum mechanical principles to explore solution spaces more efficiently than classical counterparts.

Reinforcement Learning (RL) has emerged as a promising approach for addressing these challenges. Deep Q-Networks (DQN) utilize neural networks to map state spaces to action Q-values, enabling optimal action selection \cite{Mnih2015}. The RL field has demonstrated impressive capabilities across games \cite{silver2017mastering}, continuous control \cite{lillicrap2015continuous}, locomotion \cite{peng2017deeploco}, navigation \cite{bouton2019safe}, and robotics \cite{kormushev2013reinforcement}, while showing effectiveness in solving optimization problems relevant to resource allocation \cite{Barrett2020}.
Recent quantum computing advancements have introduced new possibilities, particularly through Quantum Reinforcement Learning (QRL). Quantum computing leverages superposition and entanglement to explore solution spaces infeasible for classical computers \cite{sotelo2023quantum}. A key QRL approach utilizes Variational/Parameterized Quantum Circuits (VQCs/PQCs) \cite{jerbi2021parametrized,skolik2022quantum}, optimizable with classical ML techniques. VQCs function as quantum feature extractors, capturing complex data correlations challenging for classical models, thereby enhancing RL agent representation capabilities for improved decision-making in complex environments.

Hence, we introduce a QRL framework called the \emph{Variational Quantum Rainbow Deep Q-Network} (VQR-DQN), integrating VQC-based quantum feature extraction with advanced RL techniques. Our key contributions are:
\textbf{(1) VQR-DQN Framework:} A novel architecture combining quantum-enhanced Ring-topology VQCs with Rainbow DQN \cite{hessel2018rainbow}, incorporating distributional $Q$-learning, prioritized replay, and noisy networks.
\textbf{(2) Resource Allocation Environment:} We demonstrate the VQR-DQN effectiveness via the Human Resource Allocation Problem (HRAP) as an MDP with a comprehensive environment design simulating real-world personnel dispatch scenarios.
\textbf{(3) Experimental Evaluation:} Extensive experiments demonstrating VQR-DQN's significant outperformance over Double DQN \cite{Hasselt2016} and Rainbow DQN \cite{hessel2018rainbow} in task completion time and resource utilization across varying HRAP complexity scenarios.

\section{Related Work}
\subsection{Resource Allocation Problem}
Resource allocation problem represents a long-standing problem in operations research and management science. Traditional approaches employ mathematical optimization techniques, i.e., Linear Programming (LP) \cite{azimi2013optimal}, Mixed-Integer Linear Programming (MILP) \cite{ercsey2024multicommodity}, and branch-and-bound algorithms (B\&B) \cite{vila2014branch}, modeling allocation problems as linear equations with constraints. However, these methods are limited to small-scale problems due to exponential computational complexity growth. Heuristic and metaheuristic approaches, such as Genetic Algorithms (GA) \cite{mutlu2013iterative}, Simulated Annealing (SA) \cite{manavizadeh2013simulated}, and Particle Swarm Optimization (PSO) \cite{jarboui2008combinatorial}, provide approximate solutions for large-scale problems \cite{bouajaja2017survey}, yet suffer from domain-specific parameter tuning requirements and limited adaptability.
RL has emerged as a promising resource allocation problem paradigm. Early applications used tabular approaches like Q-learning \cite{watkins1992q} for small state-action spaces. Deep RL (DRL) advancement enabled high-dimensional problem handling through DQN and variants, i.e., DDQN \cite{Hasselt2016} and Dueling DQN \cite{wang2016dueling}, applied to this problem using neural networks for action-value function approximation \cite{paduraru2021task,du2023automatic}. Recent work \cite{nguyen2021can} has integrated RL with search-based methods like Monte Carlo tree search (MCTS) \cite{silver2017mastering}, combining RL and heuristic search strengths for complex dynamic environments. Challenges remain in scaling RL methods to real-world resource allocation problems, particularly regarding high-dimensional state representations, policy robustness, and convergence efficiency.

\subsection{Quantum Reinforcement Learning (QRL)}

QRL emergence has introduced new opportunities for complex optimization problems like HRAP. Primary advantages include handling exponentially large state-action spaces and simultaneous exploration of multiple solutions via quantum superposition and entanglement \cite{meyer2022survey,dong2008quantum,chen2020variational}. VQCs enhance agent environment understanding by extracting high-dimensional features difficult for classical methods.
Recent research explores VQCs \cite{jerbi2021parametrized,skolik2022quantum,andres2022use} as quantum feature extractors within RL frameworks, encoding classical data into quantum states and applying parameterized quantum gates to capture complex data correlations. While QRL application to resource allocation tasks remains emerging \cite{ansere2023quantum,andres2022use,xu2025quantum}, HRAP applications show significant promise. Integrating quantum computing techniques into existing RL frameworks aims to overcome classical approach limitations and achieve improved scalability and performance in real-world resource allocation tasks.
\section{Environment}\label{sec:environment}
\begin{figure*}[t]
    \centering
    \includegraphics[width=.8\linewidth]{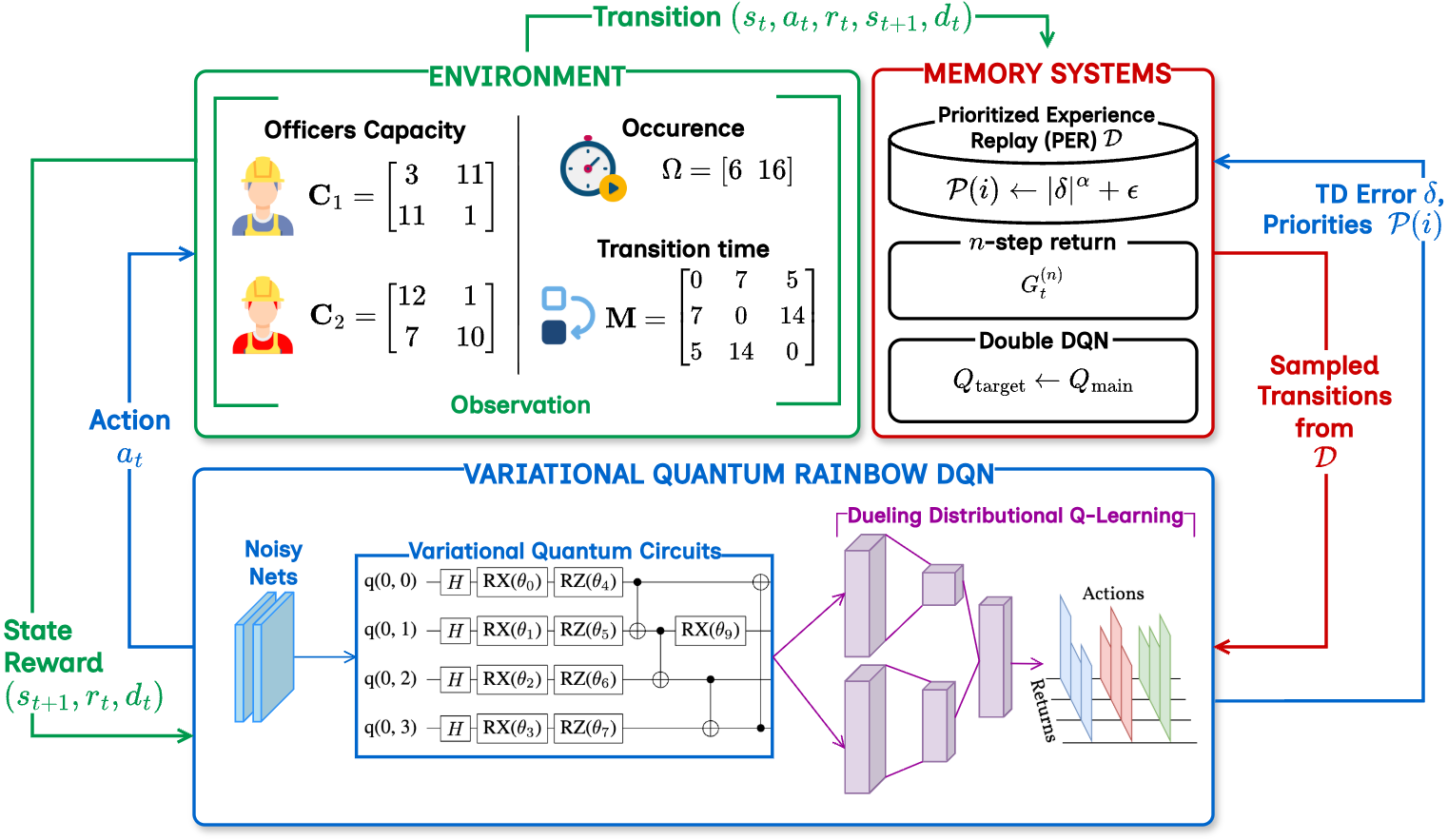}
    \caption{Architecture of VQR-DQN in solving HRAP environment. Ring-topology VQCs are integrated into the Rainbow DQN pipeline, combining noisy exploration, prioritized replay, $n$-step returns, Double DQN, and dueling distributional $Q$-learning.}
    \label{fig:hrap}
\end{figure*}

Our HRAP environment, as illustrated in Figure~\ref{fig:hrap}, is formulated as an MDP, defined by the tuple \( (\mathcal{S}, \mathcal{A}, P, R, \gamma) \), where \( \mathcal{S} \) is the state space representing the environment at each time step, \( \mathcal{A} \) is the action space, representing task assignments for officers, \( P(s' | s, a) \) is the transition probability, which defines how the environment evolves based on the agent's actions, \( R(s, a) \) is the reward function, representing the feedback signal for the agent's actions, \( \gamma \) is the discount factor, which prioritizes immediate rewards over future rewards.

In RL, the agent interacts with the environment by observing the current state, selecting an action, and receiving feedback in the form of a transition tuple:
\begin{equation}
    (s_t, a_t, r_t, s_{t+1}, d_t),
\end{equation}
where \( s_t \) is the current state at time step \( t \), \( a_t \) is the action taken by the agent in state \( s_t \), \( r_t \) is the reward received from the environment after taking action \( a_t \), \( s_{t+1} \) is the next state resulting from the action \( a_t \), \( d_t \) is the termination flag indicates whether the episode has terminated.
At each time step \( t \), the agent observes \( s_t \), selects an action \( a_t \) from the available action space, and receives a reward \( r_t \). The agent then transitions to a new state \( s_{t+1} \) and repeats the process until the episode ends, as indicated by \( d_t = 1 \). The agent aims to learn an optimal policy that maximizes the cumulative reward over time by accurately estimating the expected future rewards, or $Q$-values.

\paragraph{Entities}
Our HRAP environment consists of three primary entities: Officers, Events, and Tasks, which define the core elements of the problem: \textbf{(1) Officers:} Let $O$ denote the number of officers, where each officer $o \in \{1, \dots, O\}$ is assigned to perform specific tasks across various events. \textbf{(2) Events:} The set of scheduled events, each consisting of multiple tasks. Let  $E$ denote the number of events, indexed by $e \in \{1, \dots, E\}$, where each event requires the completion of all associated tasks within a specified timeframe. \textbf{(3) Tasks:} The set of tasks within each event. Let $T$ denote the number of tasks per event, indexed by $t \in \{1, \dots, T\}$. 

\paragraph{Objective} The objective is to assign officers to tasks in a manner that minimizes the maximum completion time across all events. The completion time for an event is determined by the time taken to complete all its tasks, considering both task execution times and transition times between events.

\subsection{State Space}

The state space \( \mathcal{S} \) represents the current configuration of the environment, containing all the necessary information for the agent to make informed decisions. In the HRAP environment, the state \( s_t \) is a high-dimensional vector comprising the following components:

\textbf{Officers' Capability Matrices} \( \mathbf{C}_o \in \mathbb{Z}^{E \times T} \) for each officer \( o \). Each entry \( C_{o,e,t} \) represents the time required for officer \( o \) to complete task \( t \) in event \( e \). The capability matrices are initialized with random integer values uniformly sampled from the interval $[1, 20]$.
  
\textbf{Event Occurrence Times} \( \Omega \in \mathbb{Z}^E \), where \( \Omega_e \) denotes the start time of event \( e \). These occurrence times are randomly generated as integers from the interval $[1, 20]$ and are sorted in ascending order to ensure temporal ordering.

\textbf{Transition Matrix} \( \mathbf{M} \in \mathbb{Z}^{(E+1) \times (E+1)} \), representing the time required for an officer to travel between events. Each entry \( M_{e_1, e_2} \) indicates the transition time from event \( e_1 \) to event \( e_2 \). The matrix is symmetric with zero diagonal entries, ensuring no time is required to remain at the same event.

The state of the environment is represented by the concatenation of the flattened capability matrices of all officers' capabilities, event occurrence times, and the transition matrix:
\begin{equation}
\begin{gathered}
\mathbf{s} = \text{Concat}\left( \text{Flatten}(\mathbf{C}_1, \dots, \mathbf{C}_N), \Omega, \text{Flatten}(\mathbf{M}) \right) \in \mathbb{R}^{d}, 
\\
d = O \times E \times T + E + (E+1)^2.
\end{gathered}
\end{equation}

\subsection{Action Space}
The action space \( \mathcal{A} \) defines the possible task assignments the agent can make at each time step $t$. An action \( a_t \) consists of assigning an officer to a task within an event. For each task \( t \) in event \( e \), the agent selects an officer \( o \) from the pool of available officers:
\begin{equation}
    a_t = \{(e, t, o) \mid e \in \{1, \dots, E\}, 
    t \in \{1, \dots, T\},
    o \in \{1, \dots, O\}\},
\end{equation}
given \( E \) events, each with \( T \) tasks, and \( O \) officers, the total number of possible actions is \( O^{E \times T} \), making the action space combinatorially large.

\subsection{Reward Function}
The reward function \( R(s, a) \) is designed to motivate efficient task allocations by minimizing the maximum time taken to complete any event. At each time step, the agent receives a reward \( r_t \) based on the negative completion time for the slowest event, normalized by the maximum possible completion time \(\Psi\). The maximum completion time \(\Psi\) is defined as:
\begin{equation}
\Psi = \left( \max(\mathbf{C}) \times E \times T \right) + \left( \max(\mathbf{M}) \times E \times T \right).
\end{equation}
The reward \( r_t \) is then calculated as:
\begin{equation}
r_t = -\frac{\max_{e} \left( \sum_{t} C_{o,e,t} + \sum_{\text{transitions}} M_{e_1, e_2} \right)}{\Psi}.
\end{equation}
This reward structure encourages the agent to minimize the longest completion time across all events, promoting efficient task assignments and travel schedules. By normalizing the reward with \(\Psi\), the reward values are scaled to a consistent range, improving the learning's stability and convergence.
\section{Variational Quantum Rainbow Deep Q-Network (VQR-DQN)}
We propose the \textit{Variational Quantum Rainbow Deep Q-Network} (VQR-DQN), a DRL framework integrating VQCs as quantum-enhanced feature extractors with Rainbow DQN mechanisms (dueling distributional $Q$-learning, noisy exploration, prioritized replay, $n$-step returns, and DDQN). Leveraging quantum-enhanced feature extraction with Ring topology, VQR-DQN captures complex correlations within high-dimensional, entangled HRAP state spaces comprising officers' capabilities, event occurrences, and transition matrices.

\subsection{Variational Quantum Circuits (VQCs)}
In our VQR-DQN framework, VQCs serve as high-dimensional, entangled feature extractors employing parameterized quantum operations in Ring topology for quantum data transformations. Detailed definitions of fundamental VQCs' terminology are provided in Appendix~\ref{app:term}.

\subsubsection{Circuit Architecture}
\begin{figure*}[hbt!]
    \centering
    \hspace{12pt} 
    \input{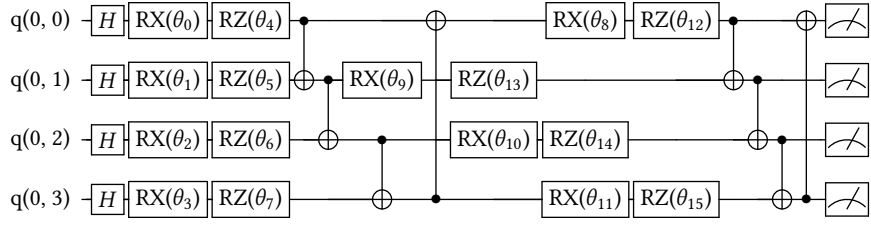}
    \caption{The visualization of the ansatz with the Ring topology with 4 input qubits and 2 layers in our designed VQC.}
    \label{fig:vqc_topology}
\end{figure*}

Our designed VQCs comprise multiple layers, each consisting of parameterized single-qubit rotations followed by entangling \(\mathrm{CNOT}\) gates arranged in a Ring topology.
Figure \ref{fig:vqc_topology} visualizes our ansatz with 4 input qubits and 2 layers.
This ansatz architecture ensures global entanglement across all qubits, boosting the capture of complex feature correlations. 

Let \(n_q\) denote the number of qubits and \(n_l\) the number of layers in the VQC. The quantum state evolves through each layer \(l \in \{1, \dots, n_l\}\) as follows:
\paragraph{Initialization}
Each qubit is initialized in the ground state \(\vert 0 \rangle\), and a layer of Hadamard gates \(H\) is applied to create an equal superposition:
\begin{equation}
    \vert \psi_0 \rangle = \bigotimes_{i=1}^{n_q} \vert 0 \rangle_i, \quad \vert \psi_{\text{init}} \rangle = H^{\otimes n_q} \vert \psi_0 \rangle, \quad H = \frac{1}{\sqrt{2}}
\begin{pmatrix}
1 & 1 \\
1 & -1
\end{pmatrix}.
\end{equation}

\paragraph{Parameterized Rotations}
For each qubit \(i\) in layer \(l\), we apply parameterized rotation gates \(\mathrm{RX}(\theta_{l,i}^{(x)})\) and \(\mathrm{RZ}(\theta_{l,i}^{(z)})\) for Pauli-X and Pauli-Z, respectively:
\begin{equation}
\begin{gathered}
    \mathrm{RX}(\theta)
    =
    e^{-i \theta X / 2} 
    =
    \begin{pmatrix}
    \cos(\tfrac{\theta}{2}) & -i \sin(\tfrac{\theta}{2}) \\
    -i\sin(\tfrac{\theta}{2}) & \cos(\tfrac{\theta}{2})
    \end{pmatrix}, \\
    \mathrm{RZ}(\phi)
    =
    e^{-i \phi Z / 2}
    =
    \begin{pmatrix}
    e^{-i\phi/2} & 0 \\
    0 & e^{i\phi/2}
    \end{pmatrix}.
\end{gathered}
\end{equation}

\paragraph{Entangling Gates}
Following the rotations, we introduce entanglement using Controlled-NOT (CNOT) gates in a Ring topology. The CNOT gate between qubits \( q_i \) (control) and \( q_{i+1} \) (target) is represented by the matrix:
\begin{equation}
    \mathrm{CNOT} = 
    \begin{pmatrix}
    1 & 0 & 0 & 0 \\
    0 & 1 & 0 & 0 \\
    0 & 0 & 0 & 1 \\
    0 & 0 & 1 & 0
    \end{pmatrix}.
\end{equation}
For \( n_q \) qubits, the entangling layer applies CNOT gates as follows:
\begin{math}
 \mathrm{CNOT}(q_i \rightarrow q_{(i+1) \bmod n_q})
\end{math}
for each qubit \(i \in \{1, \dots, n_q\}\). In total, each layer \(l\) produces a unitary \(U^{(l)}(\boldsymbol{\theta})\) comprising all single-qubit rotations and entangling gates.

\paragraph{Measurement and Final Quantum State}
For each layer \(l\), the composite unitary operation applied to all qubits is:
\begin{equation}
    U^{(l)}(\boldsymbol{\theta}^{(l)}) = \left( \bigotimes_{i=1}^{n_q} \mathrm{RX}(\theta_{l,i}^{(x)}) \cdot \mathrm{RZ}(\theta_{l,i}^{(z)}) \right) \cdot \mathrm{CNOT}_{\text{Ring}}.
\end{equation}
After \(n_l\) layers, the cumulative unitary \(U(\boldsymbol{\theta})\) is:
\begin{math}
    U(\boldsymbol{\theta}) = \prod_{l=1}^{n_l} U^{(l)}(\boldsymbol{\theta}^{(l)}),
\end{math}
Thus, the final quantum state is:
\begin{math}
\vert \psi_{\mathrm{out}}(\boldsymbol{\theta}) \rangle
\;=\;
U(\boldsymbol{\theta})
\,\vert \psi_{\mathrm{init}} \rangle
\end{math}
We measure Pauli-Z operators on each qubit:
\begin{math}
\hat{Z}_i
\;=\;
I \otimes \dots \otimes Z \otimes \dots \otimes I,
\end{math}
yielding expectation values \(\langle \psi_{\mathrm{out}} \vert \hat{Z}_i \vert \psi_{\mathrm{out}} \rangle \in [-1,1]\). Collecting them for \(i=1,\dots,n_q\) yields a quantum feature vector:
\begin{equation}
\mathbf{q}(\boldsymbol{\theta})
\;=\;
\bigl(\langle \hat{Z}_1\rangle,\dots,\langle \hat{Z}_{n_q}\rangle\bigr)^\top \in \mathbb{R}^{n_q},
\end{equation}
which we feed into subsequent classic layers in the network.

\subsection{Rainbow DQN Integration}
Our VQR-DQN incorporates all components of Rainbow DQN, augmented by the quantum feature extractor. Below, we detail each component and its role in the overall architecture.

\subsubsection{Noisy Networks for Exploration}
The VQR-DQN employs noisy dense layers to facilitate exploration by introducing trainable noise into the network parameters. This approach allows the agent to explore without relying solely on the \( \epsilon \)-greedy policy. The noisy layers modify the weights and biases as follows:
\begin{equation}
\mathbf{w} = \mathbf{w}_\mu + \mathbf{w}_\sigma \odot \boldsymbol{\epsilon}_w,
\quad
\mathbf{b} = \mathbf{b}_\mu + \mathbf{b}_\sigma \odot \boldsymbol{\epsilon}_b,
\end{equation}
where \( \mathbf{w}_\mu, \mathbf{w}_\sigma \) and \( \mathbf{b}_\mu, \mathbf{b}_\sigma \) are trainable parameters, and \( \boldsymbol{\epsilon}_w \), \( \boldsymbol{\epsilon}_b \) are noise variables sampled from a Gaussian distribution. The noise injection encourages the network to explore more diverse actions by perturbing the $Q$-values during training.

\subsubsection{Prioritized Experience Replay}
The Prioritized Experience Replay (PER) buffer enables the agent to focus more on transitions that have a higher learning potential by sampling based on the temporal-difference (TD) error \( \delta_i \). The sampling probability for each transition \( i \) is given by:
\begin{equation}
p_i = \frac{\left|\delta_i\right|^\alpha + \epsilon}{\sum_j \left|\delta_j\right|^\alpha + \epsilon}
\label{eq:sampling_prob_per},
\end{equation}
where \( \delta_i \) is the TD error for the transition is computed as:
\begin{math}
\delta_i = r_t + \gamma \max_{a'} Q_{\text{target}}(s_{t+1}, a') - Q_{\text{main}}(s_t, a_t)
\label{eq:transition_per}
\end{math},
where \( \alpha \) controls the degree of prioritization, and \( \epsilon \) ensures that all transitions have a non-zero probability of being sampled. 

\subsubsection{$n$-step Returns}
To provide richer learning signals, the VQR-DQN computes \( n \)-step returns for sampled transitions. The \( n \)-step return for a transition starting at time step \( t \) is defined as:
\begin{equation}
G_t^{(n)} = \sum_{k=0}^{n-1} \gamma^k r_{t+k} + \gamma^n Q_{\text{target}}(s_{t+n}, a_{t+n}),
\end{equation}
where \( \gamma \) is the discount factor, \( r_{t+k} \) is the reward at step \( t+k \), and \( Q_{\text{target}} \) is the target network. The \( n \)-step return combines immediate and future rewards over a longer horizon, improving the stability and efficiency of learning.

\subsubsection{Double DQN (DDQN)}
To mitigate overestimation bias, VQR-DQN employs the DDQN technique \cite{Hasselt2016}. The target $Q$-value is computed using the main network to select actions and the target network to evaluate them:
\begin{equation}
a^* = \arg\max_{a'} Q_{\text{main}}(\mathbf{s}', a'; \theta), \quad
y = r + \gamma Q_{\text{target}}(\mathbf{s}', a^*; \theta^{-}),
\end{equation}
where \( Q_{\text{main}} \) and \( Q_{\text{target}} \) denote the main and target networks, respectively.

\subsubsection{Dueling Distributional Q-Learning}
The VQR-DQN integrates the dueling architecture into the distributional $Q$-learning (C51) framework to improve stability and learning efficiency. In this approach, the $Q$-value function is decomposed into two streams: \textbf{Value Stream \( \mathbf{V}(s) \)} represents the state value, independent of actions, \textbf{Advantage Stream \( \mathbf{A}(s, a) \)} represents the relative benefit of taking action \( a \) in state \( s \).
The network outputs two sets of logits, \(\mathbf{h}_V(s)\) and \(\mathbf{h}_A(s)\), which are transformed into probability distributions using the Softmax function:
\begin{equation}
\begin{gathered}
    \mathbf{V}(s) = \text{Softmax}(\mathbf{h}_V(s)) \in \mathbb{R}^{1 \times N_{\text{atoms}}},\\
    \mathbf{A}(s) = \text{Softmax}(\mathbf{h}_A(s)) \in \mathbb{R}^{|\mathcal{A}| \times N_{\text{atoms}}}
\end{gathered}
\end{equation}
The final $Q$-value distribution \(\mathbf{p}(s, a)\) for each action is computed by combining the value and advantage streams:
\begin{equation}
  \mathbf{p}(s, a) = \mathbf{V}(s) + \left(\mathbf{A}(s, a) - \frac{1}{|\mathcal{A}|} \sum_{a'} \mathbf{A}(s, a')\right).
\end{equation}
This formulation ensures the advantage function has zero mean across actions, making the network more stable. The output \(\mathbf{p}(s, a)\) represents a categorical distribution over discrete support points (atoms), capturing the uncertainty in future rewards. 
The complete VQR-DQN framework is detailed in Algorithm~\ref{algo:vqr_dqn_dueling}, which describes the quantum-enhanced forward pass with Ring-topology VQC feature extraction and dueling distributional heads.

\begin{algorithm}
\caption{VQR-DQN Algorithm}
\label{algo:vqr_dqn_dueling}
\KwIn{State vector \( \mathbf{s} \in \mathbb{R}^{n_{\text{state}}} \)}
\KwOut{Distribution of $Q$-values \( \mathbf{p}(s, a) \) for each action \( a \in \mathcal{A} \)}

\SetKwBlock{Initialize}{Step 1: Initialize Network Parameters}{end}
\Initialize{
    Number of qubits \( n_q \), layers in VQC \( n_l \) \\
    Noisy layers with random weights \( \mathbf{w}_\mu, \mathbf{w}_\sigma \) \\
    Distributional atoms \( \mathcal{Z} = \{z_1, \dots, z_{N_{\text{atoms}}} \} \)
}

\SetKwBlock{InputProcessing}{Step 2: Input Processing}{end}
\InputProcessing{
    \( \mathbf{x} \leftarrow \mathrm{NoisyDense}(\mathbf{s}, 512, \mathrm{ReLU}) \) \\
    \( \mathbf{x} \leftarrow \mathrm{NoisyDense}(\mathbf{x}, 512, \mathrm{ReLU}) \)
}

\SetKwBlock{QuantumFeature}{Step 3: Quantum Feature Extraction using VQC}{end}
\QuantumFeature{
    \( \mathbf{q}_{\text{encoded}} \leftarrow \mathrm{Dense}(2 \times n_q \times n_l, \tanh)(\mathbf{x}) \) \\
    \For{\( l = 1 \) \textbf{to} \( n_l \)}{
        Apply Hadamard gates to all qubits: \( H^{\otimes n_q} \) \\
        Apply parameterized rotations \( \mathrm{RX}(\theta_{l,i}^{(x)}) \) and \( \mathrm{RZ}(\theta_{l,i}^{(z)}) \) to each qubit \( i \) \\
        Apply Ring entanglement: \( \mathrm{CNOT}(q_i \rightarrow q_{(i+1) \bmod n_q}) \) \\
    }
    Measure Pauli-Z expectation values: \( \mathbf{q}(\boldsymbol{\theta}) \leftarrow [\langle \hat{Z}_1 \rangle, \dots, \langle \hat{Z}_{n_q} \rangle] \)
}

\SetKwBlock{DuelingHead}{Step 4: Dueling Distributional Head}{end}
\DuelingHead{
    Value stream: \( \mathbf{V}(s) \leftarrow \mathrm{Softmax}(\mathbf{h}_V(s)) \) \\
    Advantage stream: \( \mathbf{A}(s, a) \leftarrow \mathrm{Softmax}(\mathbf{h}_A(s, a)) \) \\
    Final $Q$-value distribution: \( \mathbf{p}(s, a) = \mathbf{V}(s) + \left(\mathbf{A}(s, a) - \frac{1}{|\mathcal{A}|} \sum_{a'} \mathbf{A}(s, a') \right) \)
}

\Return{Distribution \( \mathbf{p}(s, a) \forall a \in \mathcal{A} \)}
\end{algorithm}

\section{Implementation}
This section describes the training procedure of the VQR-DQN framework, including network initialization, agent-environment interaction, and network optimization. Algorithm~\ref{algo:vqr_dqn_training} outlines the training procedure incorporating prioritized replay, $n$-step returns, and target network updates.

\subsection{Network Initialization}
The VQR-DQN agent consists of a main network \( Q_{\text{main}} \) and a target network \( Q_{\text{target}} \), both initialized with the same parameters at the start of training. The network comprises noisy dense layers followed by a VQC for feature extraction and a dueling distributional head to estimate the $Q$-value distribution for each action. The target network is updated periodically to stabilize the learning process.
The replay buffer \( \mathcal{D} \) is initialized as a PER buffer, which allows the agent to sample transitions based on their TD errors. The exploration strategy begins with a high exploration rate \( \epsilon \) and gradually decays over time using an \(\epsilon\)-greedy policy.

\subsection{Agent-Environment Interaction}
\paragraph{Action Selection.} Each episode resets environment to initial state \( s_0 \). At time step \( t \), agents select action \( a_t \) using \( \epsilon \)-greedy policy: with probability \( \epsilon \), random action from \( \mathcal{A} \); otherwise, action maximizing $Q$-value from main network \( Q_{\text{main}}(s_t, a) \). Transition tuples \( (s_t, a_t, r_t, s_{t+1}, d_t) \) store in replay buffer, where \( d_t \) indicates episode termination.

\paragraph{Exploration-Exploitation Strategy.} Exploration rate \( \epsilon \) follows exponential decay: $\epsilon \leftarrow \max(\epsilon \times \epsilon_{\text{decay}}, \epsilon_{\text{min}})$, where \( \epsilon_{\text{decay}} \) controls decay rate and \( \epsilon_{\text{min}} \) sets lower bound.

\subsection{Network Optimization}
During each training step, a mini-batch \( \mathcal{B} \) of transitions is sampled from the replay buffer \( \mathcal{D} \), with sampling probability proportional to the priorities (Eq. \ref{eq:sampling_prob_per}). 
For each sampled mini-batch, the agent performs a gradient descent step to minimize the loss between the predicted $Q$-value distribution and the target distribution. Here, the \( n \)-step return \( G_t^{(n)} \) is used to compute a more stable target:
\begin{equation}
G_t^{(n)} = \sum_{k=0}^{n-1} \gamma^k r_{t+k} + \gamma^n Q_{\text{target}}(s_{t+n}, a^*),
\end{equation}
where \( a^* = \arg\max_{a} Q_{\text{main}}(s_{t+n}, a) \). This \( n \)-step return provides richer learning signals by incorporating future rewards over a longer horizon.

The loss function is based on the cross-entropy between the predicted distribution and the projected target distribution:
\begin{math} \mathcal{L} = \mathbb{E}_{\mathcal{B}} \left[\delta^2\right] \end{math},
where \( \delta \) is the TD error for each transition in the mini-batch. The loss ensures that the $Q$-value estimates from the main network \( Q_{\text{main}} \) converge toward the \( n \)-step return target. 
Then, the priorities in the replay buffer are updated based on the new TD errors:
\begin{math}
\mathcal{P}(i) \leftarrow \left|\delta_i\right|^\alpha + \epsilon
\end{math},
where \( \alpha \) controls the prioritization degree, and \( \epsilon \) ensures that all transitions have non-zero priority. Gradients are clipped to prevent exploding gradients.
The parameters of the main network \( \theta \) are then updated using the Adam optimizer with a learning rate \( \eta \).
Finally, the target network \( Q_{\text{target}} \) is periodically synchronized with the main network \( Q_{\text{main}} \) to stabilize the training.

\begin{algorithm}[h]
\caption{VQR-DQN Initialization and Training Procedure}
\label{algo:vqr_dqn_training}
\KwIn{Replay buffer \(\mathcal{D}\), target network \(Q_{\text{target}}\), main network \(Q_{\text{main}}\)}
\KwOut{Updated network parameters \(\theta\)}
\SetKwBlock{Initialize}{Initialize}{end}
\Initialize{
    Replay buffer \(\mathcal{D} \leftarrow \emptyset\) \\
    Target network \(Q_{\text{target}} \leftarrow Q_{\text{main}}\) \\
    Exploration rate \(\epsilon \leftarrow 1.0\)
}

\SetKwBlock{Episode}{For each episode}{end}
\Episode{
    Reset environment: \( s_0 \leftarrow \text{env.reset()} \) \\
    \ForEach{time step \( t \)}{
        \SetKwBlock{SelectAction}{Step 1: Action Selection}{end}
        \SelectAction{
            \eIf{\( \text{random}() < \epsilon \)}{
                Select random action \( a_t \in \mathcal{A} \)
            }{
                Compute $Q$-values from the main network:
                \( Q(s_t, a) = \sum_{i=1}^{N_{\text{atoms}}} z_i \cdot p_i(s_t, a) \) \\
                Select action \( a_t = \arg\max_{a} Q(s_t, a) \)
            }
        }
        Execute action \( a_t \), observe reward \( r_t \) and next state \( s_{t+1} \) \\
        Store transition \( (s_t, a_t, r_t, s_{t+1}, d_t) \) in replay buffer \(\mathcal{D}\) \\
        Update exploration rate: \( \epsilon \leftarrow \max(\epsilon \times \epsilon_{\text{decay}}, \epsilon_{\text{min}}) \)
    }
}

\SetKwBlock{TrainingStep}{Step 2: Network Optimization}{end}
\TrainingStep{
    Sample mini-batch \( \mathcal{B} \) from prioritized replay buffer \(\mathcal{D}\) \\
    \ForEach{transition \( (s_t, a_t, r_t, s_{t+1}, d_t) \in \mathcal{B} \)}{
        Compute \( n \)-step return: 
        $G_t^{(n)} = \sum_{k=0}^{n-1} \gamma^k r_{t+k} + \gamma^n Q_{\text{target}}(s_{t+n}, a^*)$\\
        where $a^* = \arg\max_{a} Q_{\text{main}}(s_{t+n}, a)$  \\
        Compute TD error: 
        $\delta = G_t^{(n)} - Q_{\text{main}}(s_t, a_t)$\\
        Update priorities: \(\mathcal{P}(i) \leftarrow |\delta|^\alpha + \epsilon \) \\
        Perform gradient descent step using loss: 
        $\mathcal{L} = \mathbb{E}_{\mathcal{B}} \left[\delta^2\right]$\\
    }
    Periodically update target network: \( Q_{\text{target}} \leftarrow Q_{\text{main}} \)
}

\Return{Updated network parameters \(\theta\)}
\end{algorithm}
\section{Results}
We present VQR-DQN experimental setup, performance evaluation, and learning behavior analysis versus baseline (random assignment), DDQN, and Rainbow DQN using average rewards and normalized makespan reduction across varying HRAP complexity, assessing learning efficiency, stability, and final performance advantages.

\subsection{Experimental Setup}
Four HRAP configurations evaluate performance under varying complexity: 3 Officers - 2 Tasks - 2 Events ($\texttt{3O-2T-2E}$), 4 Officers - 3 Tasks - 2 Events ($\texttt{4O-3T-2E}$), 4 Officers - 3 Tasks - 3 Events ($\texttt{4O-3T-3E}$), and 5 Officers - 4 Tasks - 4 Events ($\texttt{5O-4T-4E}$). All methods were trained for 50,000 episodes under identical conditions, with the best-performing checkpoints evaluated across 200 testing episodes. Quantum simulations used TensorFlow Quantum with computations performed on IonQ Aria-1 quantum processing unit via IonQ quantum computing service.

\subsubsection{Learning Curves}
\begin{figure*}[t]
    \centering
    \begin{subfigure}[b]{0.24\textwidth}
        \centering
        \includegraphics[width=\linewidth]{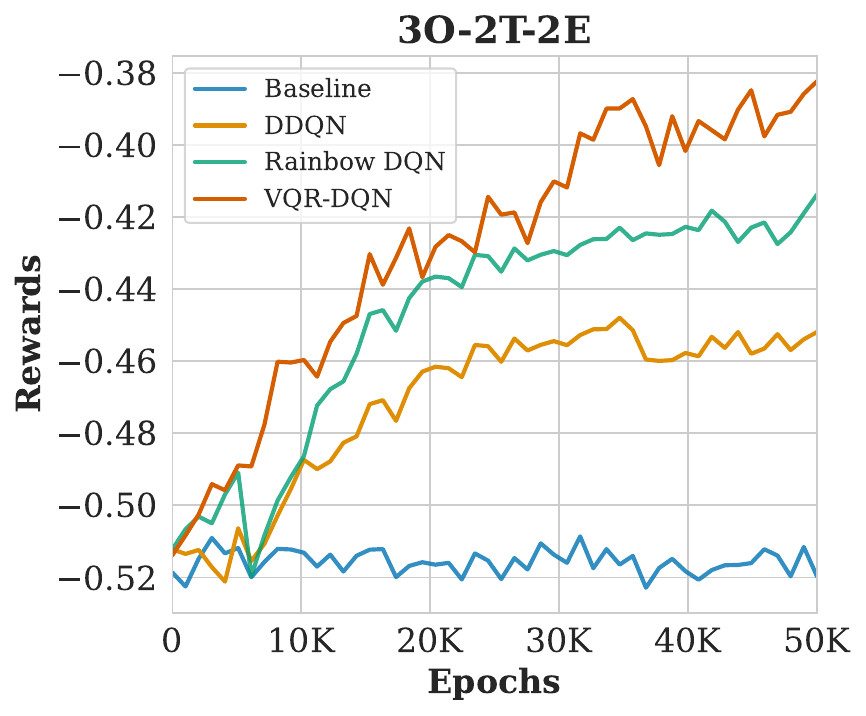}
        \caption{$|\mathcal{A}|=3^4$}
    \end{subfigure}
    \begin{subfigure}[b]{0.24\textwidth}
        \centering
        \includegraphics[width=\linewidth]{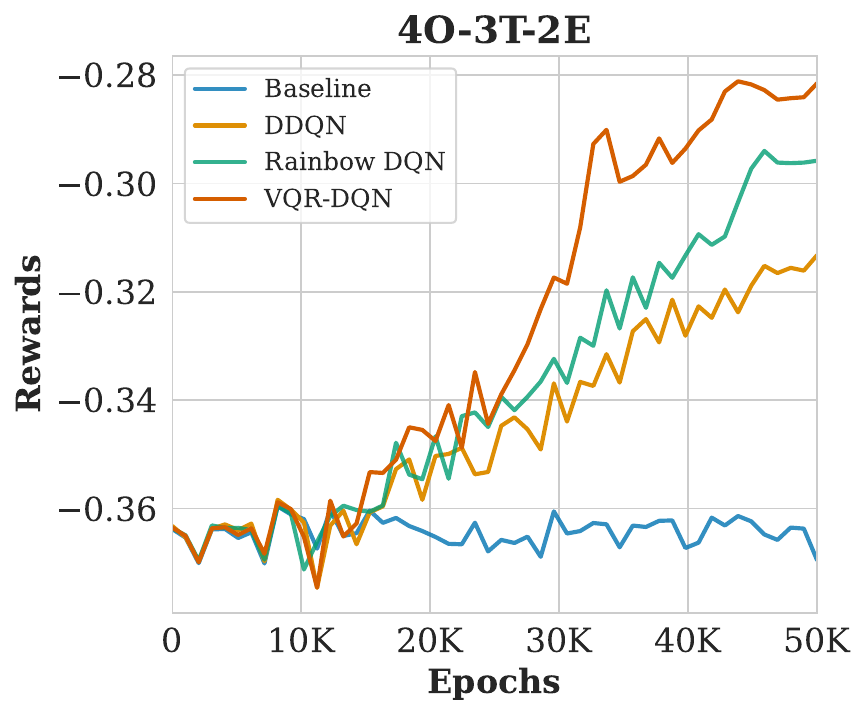}
        \caption{$|\mathcal{A}|=4^6$}
    \end{subfigure}
    \begin{subfigure}[b]{0.24\textwidth}
        \centering
        \includegraphics[width=\linewidth]{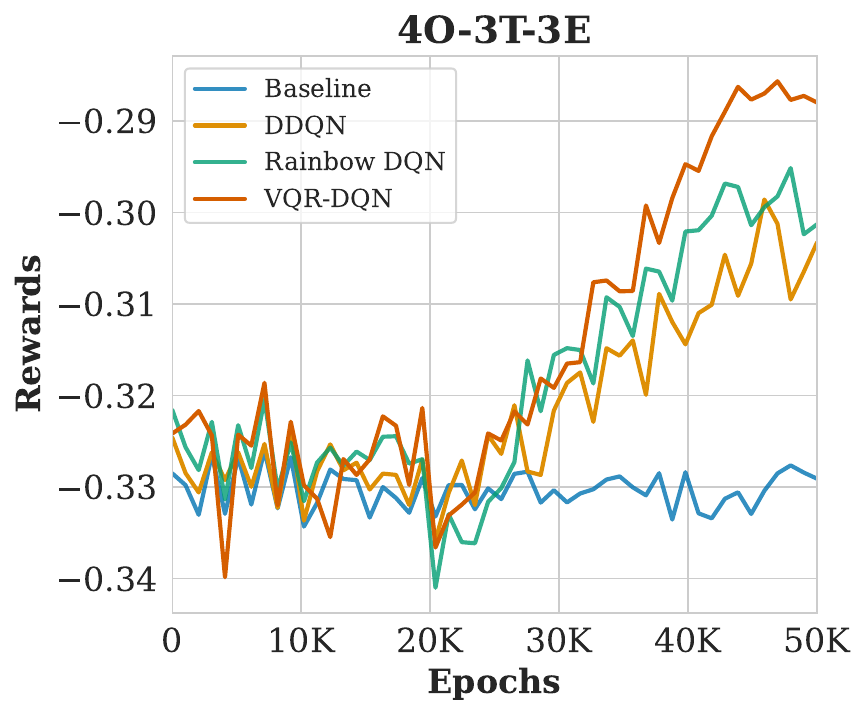}
        \caption{$|\mathcal{A}|=4^9$}
    \end{subfigure}
    \begin{subfigure}[b]{0.24\textwidth}
        \centering
        \includegraphics[width=\linewidth]{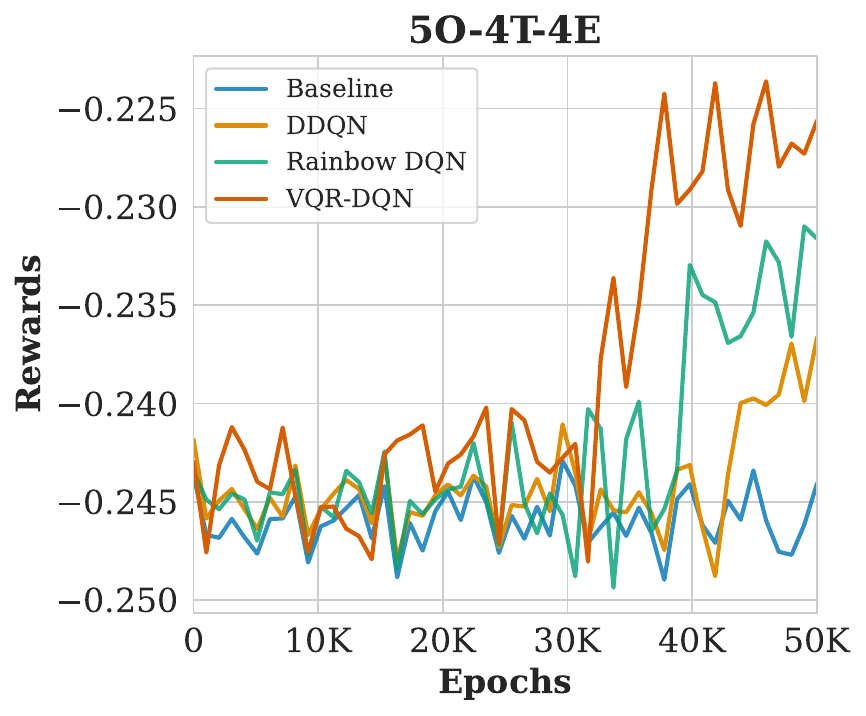}
        \caption{$|\mathcal{A}|=5^{16}$}
    \end{subfigure}
    \caption{The learning curves for VQR-DQN and other algorithms in 50,000 episodes for different HRAP configurations (with their action space).}
    \label{fig:learning_curves}
\end{figure*}

Figure \ref{fig:learning_curves} illustrates the learning curves over 50,000 training episodes for all configurations, demonstrating how each algorithm's performance evolves during the training process. 
The learning curves reveal that VQR-DQN generally achieved faster convergence and more stable learning compared to other approaches, particularly in the early stages of training. 
Learning speed and stability also vary with problem complexity. In simpler scenarios like $\texttt{3O-2T-2E}$, all algorithms achieve relatively quick and smooth convergence within the first 15,000 episodes. However, as the configuration complexity increases, particularly in $\texttt{5O-4T-4E}$, the convergence becomes notably slower and more gradual, with algorithms requiring nearly 30,000 to 40,000 episodes to stabilize. Despite this, VQR-DQN maintains more stable improvement compared to other algorithms, especially in later training stages.

\begin{table*}[t]
  \centering
  \noindent
  \begin{minipage}[t]{0.6\textwidth}
    \centering
    \small
    \captionof{table}{Averaged rewards (Normalized makespan reduction\%) over the baseline for VQR-DQN and other algorithms across 200 testing episodes, evaluated from the agent checkpoint with the highest training score. The best results are in \textbf{bold}.}
    \label{tab:test_results}
    \begin{tabular}{l|l|l|lll}
      \toprule
      Config.  & $|\mathcal{A}|$ &  Baseline & DDQN & Rainbow DQN & VQR-DQN (Ours)\\
      \midrule
      $\texttt{3O-2T-2E}$  & $3^4$ & -0.5225 & -0.4539 {\color[rgb]{0,0.5,0}(\textbf{\(\blacktriangle\)} 13.1\%)} & -0.4189 {\color[rgb]{0,0.5,0}(\textbf{\(\blacktriangle\)} 19.8\%)} & \textbf{-0.3823 {\color[rgb]{0,0.5,0}(\textbf{\(\blacktriangle\)} 26.8\%)}} \\
      $\texttt{4O-3T-2E}$  & $4^6$ & -0.3689 & -0.3132 {\color[rgb]{0,0.5,0}(\textbf{\(\blacktriangle\)} 15.1\%)} & -0.2957 {\color[rgb]{0,0.5,0}(\textbf{\(\blacktriangle\)} 19.8\%)} & \textbf{-0.2815 {\color[rgb]{0,0.5,0}(\textbf{\(\blacktriangle\)} 23.7\%)}} \\
      $\texttt{4O-3T-3E}$  & $4^9$ & -0.3316 & -0.3032 {\color[rgb]{0,0.5,0}(\textbf{\(\blacktriangle\)} 8.6\%)} & -0.3012 {\color[rgb]{0,0.5,0}(\textbf{\(\blacktriangle\)} 9.2\%)} & \textbf{-0.2872 {\color[rgb]{0,0.5,0}(\textbf{\(\blacktriangle\)} 13.4\%)}} \\
      $\texttt{5O-4T-4E}$  & $5^{16}$ & -0.2488 & -0.2366 {\color[rgb]{0,0.5,0}(\textbf{\(\blacktriangle\)} 4.9\%)} & -0.2309 {\color[rgb]{0,0.5,0}(\textbf{\(\blacktriangle\)} 7.2\%)} & \textbf{-0.2236 {\color[rgb]{0,0.5,0}(\textbf{\(\blacktriangle\)} 10.1\%)}} \\
      \bottomrule
    \end{tabular}
  \end{minipage}
  \hfill
  \begin{minipage}[t]{0.39\textwidth}
    \centering
    \small
    \captionof{table}{Averaged rewards (Normalized makespan reduction\%) over the baseline for different topologies for VQR-DQN across 200 testing episodes in the configuration $\texttt{3O-2T-2E}$ ($|\mathcal{A}|=3^4$), evaluated from the agent checkpoint with the highest training score. The best results are in \textbf{bold}.}
    \label{tab:topo_results}
    \begin{tabular}{ll}
      \toprule
      \textbf{Algorithm} & \textbf{Rewards} \\
      \midrule\midrule
      Baseline & -0.5225 \\
      \midrule
      VQR-DQN + Linear & -0.4249 {\color[rgb]{0,0.5,0}(\textbf{\(\blacktriangle\)} 18.7\%)} \\
      VQR-DQN + Star &  -0.4514 {\color[rgb]{0,0.5,0}(\textbf{\(\blacktriangle\)} 13.6\%)} \\
      VQR-DQN + Ring & \textbf{-0.3823 {\color[rgb]{0,0.5,0}(\textbf{\(\blacktriangle\)} 26.8\%)}} \\
      VQR-DQN + All-to-All & -0.4103 {\color[rgb]{0,0.5,0}(\textbf{\(\blacktriangle\)} 21.5\%)} \\
      \bottomrule
    \end{tabular}
  \end{minipage}
\end{table*}

\subsubsection{Performance Evaluation}
Table \ref{tab:test_results} shows the averaged rewards across 200 test episodes using the best checkpoint from training.
In the simplest configuration ($\texttt{3O-2T-2E}$), VQR-DQN achieved the most substantial improvement, showing a 26.8\% increase in performance compared to the baseline, while DDQN and Rainbow DQN showed improvements of 13.1\% and 19.8\% respectively. This significant enhancement suggests that the quantum-enhanced feature extraction is particularly effective in capturing important patterns in simpler action spaces.

As the problem complexity increased in the $\texttt{4O-3T-2E}$ configuration, VQR-DQN maintained its superior performance with a 23.7\% improvement over the baseline, compared to DDQN's 15.1\% and Rainbow DQN's 19.8\%. The learning curves show that VQR-DQN not only achieved better final performance but also demonstrated more stable learning progression throughout the training process.

In more complex scenarios ($\texttt{4O-3T-3E}$ and $\texttt{5O-4T-4E}$), while the relative improvements were smaller due to increased problem difficulty, VQR-DQN still maintained its advantage. For the most complex configuration ($\texttt{5O-4T-4E}$), VQR-DQN achieved a 10.1\% improvement over the baseline, outperforming both DDQN (4.9\%) and Rainbow DQN (7.2\%). VQR-DQN consistently outperformed baseline random assignment, DDQN, and Rainbow DQN across all configurations, demonstrating that quantum-enhanced feature extraction provides robust, scalable advantages for resource allocation optimization, particularly in moderate complexity scenarios.
\section{Impact of Topologies in VQCs for RL}

We examine the correlation between expressibility and VQC topology performance in RL, specifically HRAP tasks. \textbf{\textit{Expressibility}} quantifies how uniformly a circuit's random parameterizations explore state space through Kullback–Leibler divergence between circuit-produced states and uniform Haar-random distribution \cite{correr2024characterizing,druagan2022quantum}. \textbf{\textit{Entanglement}} captures average circuit entanglement generation, measured via Meyer-Wallach (MW) measure \cite{meyer2002global}, i.e., a global metric based on single-qubit reduced state purity ranging from 0 (no entanglement) to 1 (maximal multi-qubit entanglement). Alternative measures include Scott's multipartite quantifiers \cite{scott2004multipartite}, averaging bipartite entanglement across all partitions.

Recent studies systematically evaluate entangler connectivities (Ring, Linear, Star, All-to-All) using these metrics. Both theoretical analyses and our experimental results indicate that Ring topologies provide rich expressibility and entanglement structure. In our HRAP environment, VQR-DQN with Ring topology achieved superior average rewards (Ring $>$ All-to-All $\approx$ Linear $\gg$ Star) (see Table \ref{tab:topo_results}), aligning with \cite{correr2024characterizing} findings that Ring circuits exhibit the highest average expressibility and entanglement for given qubits and layers. This correlates with \cite{hubregtsen2021evaluation} observations of moderate-to-strong correlation between circuit expressibility and classification accuracy, and \cite{sim2019expressibility} noted ``substantial improvement'' in expressibility when two-qubit gates are arranged in Ring or All-to-All versus Linear topology.

Practically, Star topology entangles peripheral qubits only via the central qubit, creating a correlated entanglement structure, while Ring distributes entangling operations around the loop, promoting global entanglement \cite{correr2024characterizing}. Ring's greater entangling reach yields higher MW scores and uniform state coverage, enabling quantum agents to represent complex HRAP policies. RL often requires modeling sequential or spatial correlations; Ring's circular connectivity aligns with these structures, enabling efficient information flow and long-range dependency capture. However, the empirical connection between expressibility/entanglement metrics and VQC performance in RL tasks remains to be proven \cite{sim2019expressibility,druagan2022quantum}.

\section{Conclusion}
We introduce VQR-DQN, a hybrid quantum-classical RL agent integrating Ring-topology VQCs into Rainbow DQN. VQR-DQN achieved superior HRAP performance, reducing normalized makespan by 26.8\% versus random baseline and outperforming Double DQN and Rainbow DQN by 4.9-13.4\%. Ring-connected circuits consistently surpassed other topologies through broader entangling reach, aligning with theoretical expressibility and learning dynamics insights. These findings highlight quantum-enhanced feature extraction potential in complex decision-making, suggesting co-designed quantum architectures with RL objectives could yield further gains as quantum hardware advances.

\section*{Acknowledgement}
This work has been supported by NSERC Discovery Grant No RGPIN 2025-00129.

\printbibliography

\appendix
\section{Appendix}\label{app:term}
\begin{figure}[h]
    \centering
    \includegraphics[width=\linewidth]{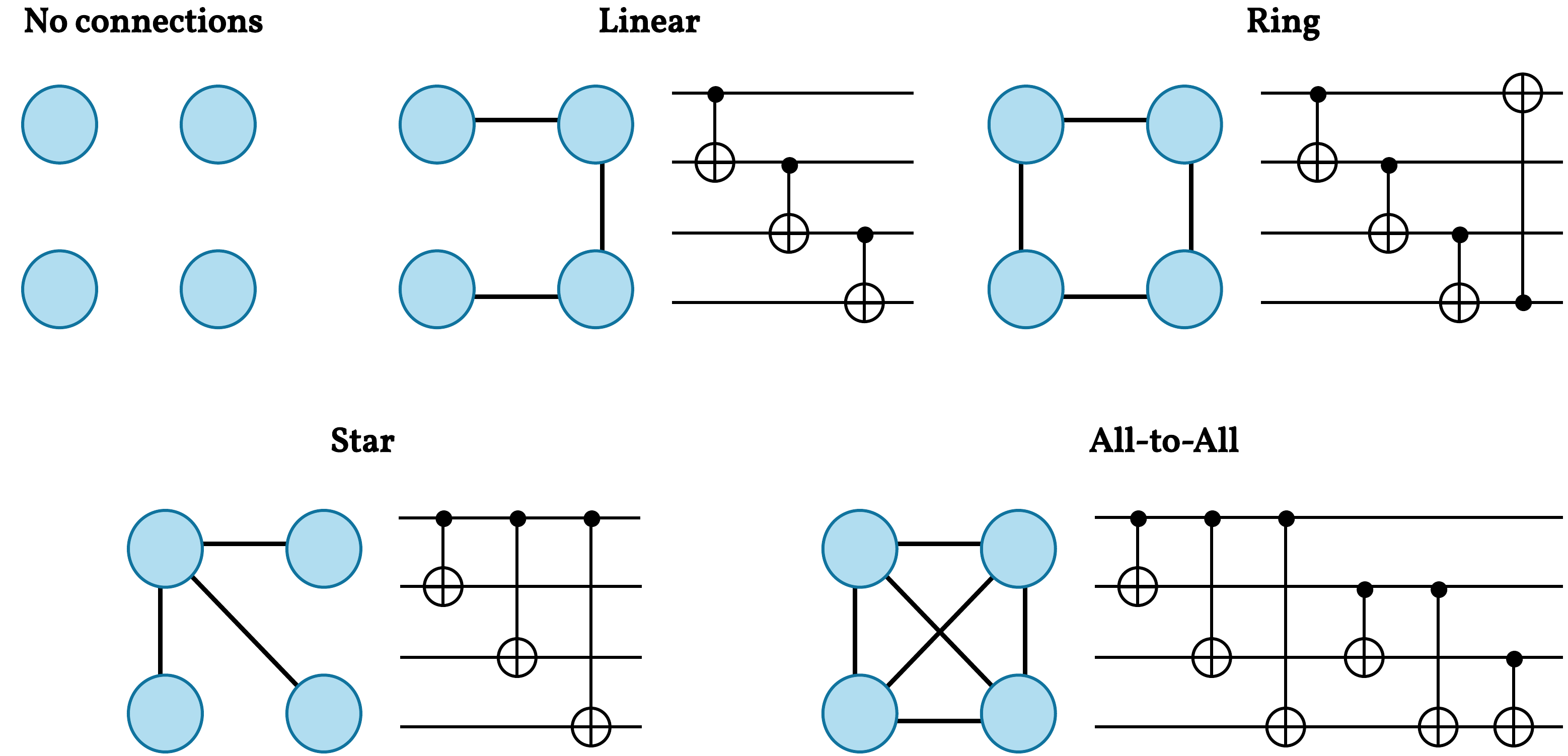}
    \caption{Graphs of the topologies observed in different quantum computer architectures.}
    \label{fig:quantum_topo}
\end{figure}
In this section, we introduce fundamental terminologies essential for understanding VQCs.
\paragraph{Topologies} (Figure \ref{fig:quantum_topo}) to the different graph topologies related to each of the connectivities between qubits that can be performed in quantum hardware. Common topologies include: 
\begin{itemize}
    \item \textbf{Linear:} Qubits lie in a chain, each entangled only with immediate neighbors.
    \item \textbf{Ring:} Arranged in a circle, each qubit is entangled with two neighbors.
    \item \textbf{Star:} A central qubit directly connects to all peripheral qubits, and the latter do not interconnect.
    \item \textbf{All-to-All:} Every qubit can entangle with every other qubit.
\end{itemize}

\paragraph{Ansätze} refers to the specific circuit structure implemented within a given topology:
\begin{itemize}
    \item \textbf{Parameterized gates:} Rotational operations ($\mathrm{RX}$, $\mathrm{RY}$, $\mathrm{RZ}$) with trainable angles that can be optimized through classical algorithms.
    \item \textbf{Entangling gates:} Fixed operations (typically $\mathrm{CNOT}$) that create quantum correlations between qubits according to the chosen topology.
    \item \textbf{Layered structure:} Repeating blocks of parameterized and entangling gates that increase circuit depth and expressivity.
    \item \textbf{Hardware-efficient ansatz:} Circuit designs that minimize the number of gates while maintaining computational power.
\end{itemize}

\end{document}